\documentclass{article}

\usepackage{microtype}
\usepackage{graphicx}
\usepackage{subfigure}
\usepackage{booktabs} 
\usepackage{amsmath}
\usepackage{amssymb}
\usepackage{amssymb,centernot}

\usepackage{hyperref}

\usepackage[accepted]{icml2020}
\icmltitlerunning{Conditioning of Reinforcement Learning Agents and its Policy Regularization Application}
\begin{document}

\twocolumn[
\icmltitle{Conditioning of Reinforcement Learning Agents and its \\ Policy Regularization Application}
\icmlsetsymbol{equal}{*}

\begin{icmlauthorlist}
\icmlauthor{Arip Asadulaev}{itmo}
\icmlauthor{Igor Kuznetsov}{itmo}
\icmlauthor{Gideon Stein}{itmo}
\icmlauthor{Andrey Filchenkov}{itmo}
\end{icmlauthorlist}

\icmlaffiliation{itmo}{ITMO University, Saint-Petersburg, Russia}

\icmlcorrespondingauthor{Arip Asadulaev}{aripasadulaev@itmo.ru}

\icmlkeywords{Machine Learning, ICML}

\vskip 0.3in
]

\makeatletter
\setlength{\@fptop}{0pt}
\makeatother
\printAffiliationsAndNotice{}
\begin{abstract}

The outcome of Jacobian singular values regularization was studied for supervised learning problems~\cite{DBLP:journals/corr/abs-1711-04735}. It also was shown that Jacobian conditioning regularization can help to avoid the ``mode-collapse'' problem in Generative Adversarial Networks~\cite{pmlr-v80-odena18a}. In this paper, we try to answer the following question: \textbf{Can information about policy conditioning help to shape a more stable and general policy of reinforcement learning agents?} To answer this question, we conduct a study of Jacobian conditioning behavior during policy optimization. To the best of our knowledge, this is the first work that research condition number in reinforcement learning agents. We propose a conditioning regularization algorithm and test its performance on the range of continuous control tasks. Finally, we compare algorithms on the CoinRun~\cite{DBLP:conf/icml/CobbeKHKS19} environment with separated train end test levels to analyze how conditioning regularization contributes to agents' generalization.

\end{abstract}

\section{Introduction}
Generalization in Reinforcement Learning~(RL) is different from supervised learning generalization problem~\cite{DBLP:journals/corr/abs-1804-06893}. We need specific techniques to avoid overfitting of RL algorithms~\cite{DBLP:journals/corr/abs-1810-00123}. Agents can achieve different scores on the test set while all of them achieved the same rewards during training. In RL, the test data performance depends on agent architecture, because different architectures have different \textit{priori} algorithmic preferences~(inductive biases)~\cite{DBLP:journals/corr/abs-1804-06893}. 

For example, Convolutional Neural Networks~(CNNs) agents are too sensitive to small visual changes and can completely fail due to perturbations~\cite{DBLP:conf/iclr/LeeLSL20}. Such techniques as the first CNNs layer randomization can avoid it and help to learn robust representations~\cite{DBLP:conf/iclr/LeeLSL20}. \textbf{In our paper, to control agents sensitivity to small changes in the environment, we propose to use agent Jacobian condition number regularization.} 

%\textbf{Conditioning} or \textbf{condition number} is the measure that indicates how much the function output can change for a small change in the input. %\textbf{Dynamical Isometry} is a neural network property stating that the distance between a network's inputs is the same as the distance between outputs.

%\textbf{Related Work:} 
For classification problems, was shown that achieving \textbf{Dynamical Isometry} property for neural networks could significantly speed up training. This property can be achieved by having a mean squared singular value equal to $\mathcal{O}(1)$ of a Jacobian input-output network~\cite{DBLP:journals/corr/abs-1711-04735}. %the role of conditioning was previously studied. %~\cite{batchnorm, rescon}. %and it was shown that well-conditioned neural networks with orthogonal weight initialization could significantly speed up training~\cite{batchnorm, rescon}. 

The role of mean squared singular values of input-output Jacobian was also studied for Generative Adversarial Networks (GANs). It was shown that conditioning of generator Jacobian is causally related to the generator performance, and a conditioning regularization can help to avoid the ``mode-collapse'' problem~\cite{pmlr-v80-odena18a}. Where \textbf{conditioning} or \textbf{condition number} is the measure that indicates how much the function output can change for a small change in the input.% (Where GANs only represent a few modes of the true distribution).
\begin{figure*}[h!]
    \centering
    \includegraphics[width=\linewidth]{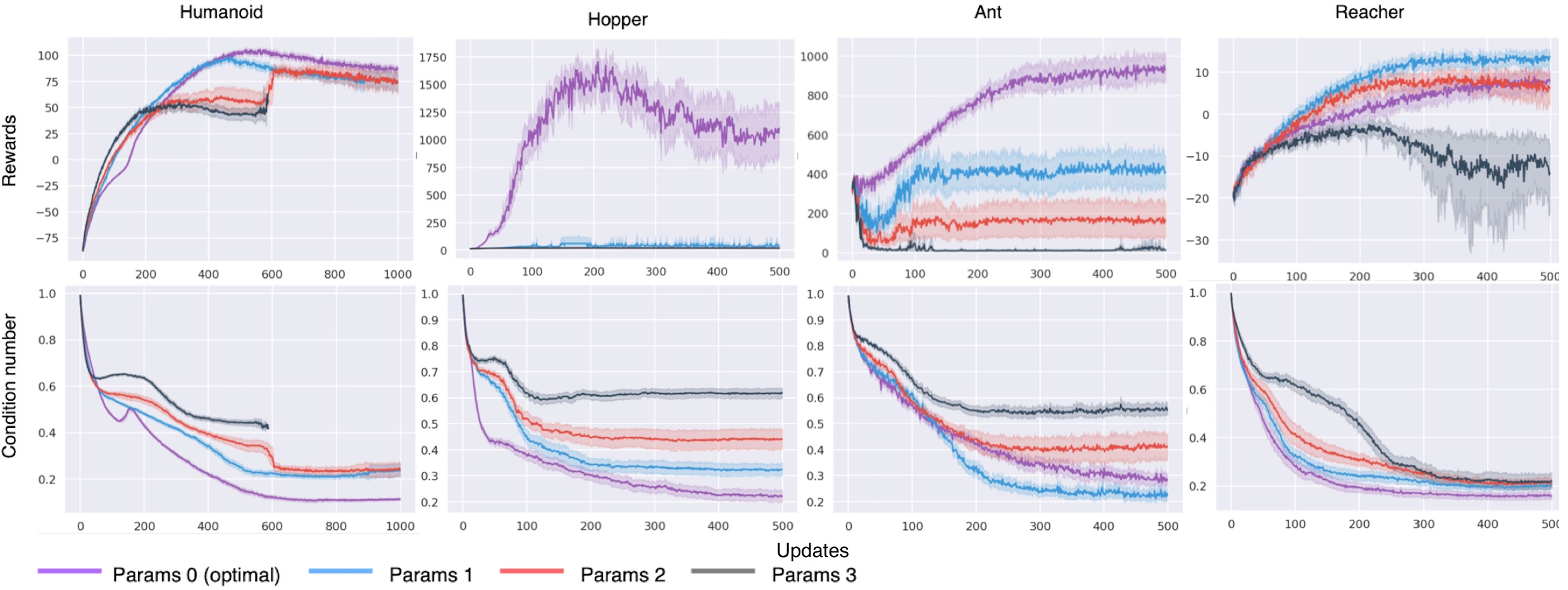}
    \vspace{-17pt}
    \caption{PPO rewards and conditioning $\psi$ in PyBullet environments with different hyperparameters. Each curve is obtained by averaging results of 30 agents (10 for each seed)}
    \label{fig:1}
    \vspace{-12.5pt}
\end{figure*}
\begin{figure*}[h!]
    \centering
    \includegraphics[width=\linewidth]{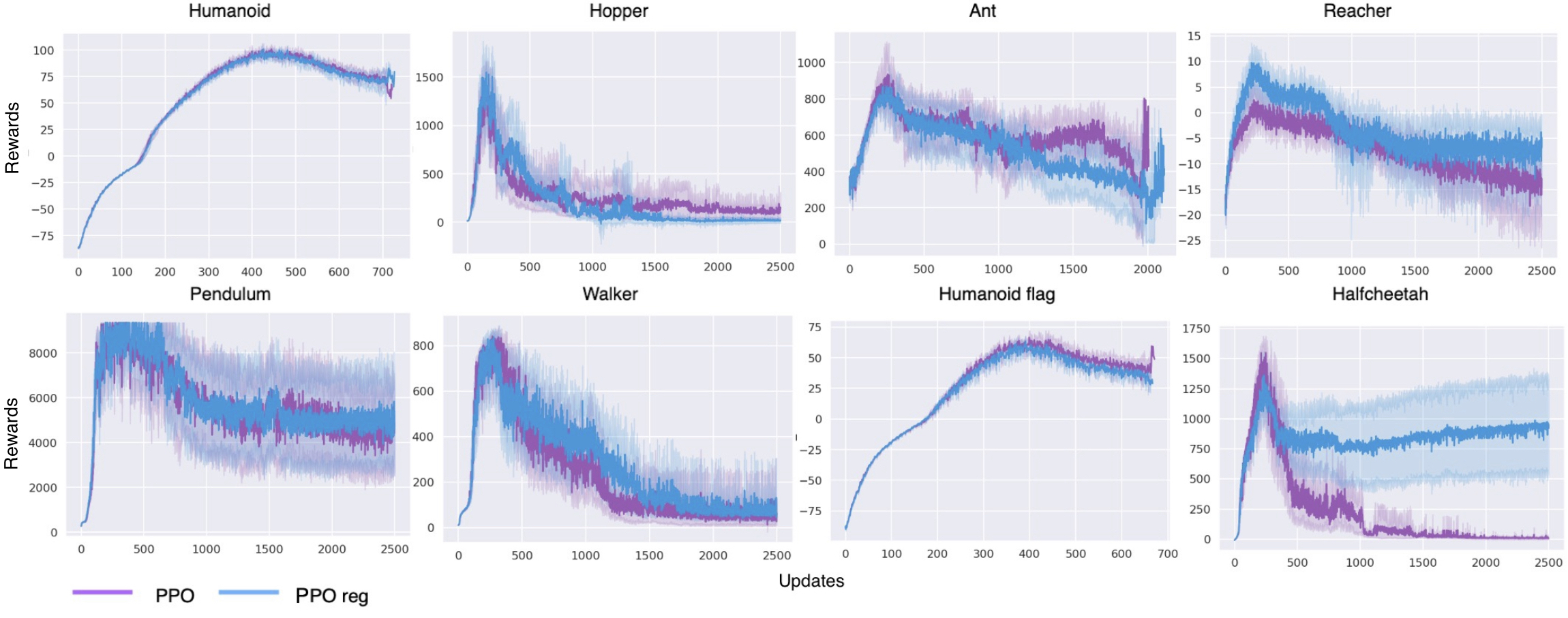}
    \vspace{-17pt}
    \caption{PPO rewards and conditioning $\psi$, in PyBullet environments. Each curve is obtained by averaging results of 30 agents (10 for each seed).}
    \label{fig:2}
    \vspace{-13pt}
\end{figure*}

%\textbf{Conditioning and Policy Performance:} 
Before using Jacobian conditioning regularization in RL agents, we conduct a study of the relationship between policy performance and Jacobian conditioning to find justifications for using it as a regularization. We analyze the behavior of the agent conditioning on different policies that are set by different sets of hyperparameters and see a correspondence between the conditioning and the ratio of achieved rewards.

Based on these observations, \textbf{we apply condition number regularization} to Trust Region Policy Optimization (TRPO)~\cite{DBLP:journals/corr/SchulmanLMJA15} and Proximal Policy Optimization~(PPO)~\cite{DBLP:journals/corr/SchulmanWDRK17} algorithms and compare their performance on 8 continuous control tasks in the PyBullet environment~\cite{pybulletgym}. %Due to impossibility of apparent exploration of all possible states in continuous state tasks, better generalization is necessary to achieve more rewards. 
In our experiments, models with regularization outperformed other models on most of the tasks. 

To explicitly test how our regularization affects on agent generalization, we run the PPO algorithm with conditioning regularization on CoinRun environments\cite{DBLP:conf/icml/CobbeKHKS19}. The results are presented in Section 3.

\section{Conditioning of RL Agents}
\textbf{Conditioning Estimation:}  Getting the mean squared singular value of a Jacobian input-output network using Singular Value Decomposition~(SVD) is time-consuming. Because of this, we adapted a technique designed to assess GAN models conditioning to RL agents for fast conditioning estimation. Jacobian Clamping (JC)~\cite{pmlr-v80-odena18a} is an algorithm that computes the condition number of the generator's Jacobian and locks it inside the interval where the Dynamical Isometry property can be achieved. %We use this algorithm for fast condition number estimation in RL agents. Empirical studies are performed in order to verify how condition number relates to policy performance.

\begin{figure*}[h!]
    \centering
    \includegraphics[width=\linewidth]{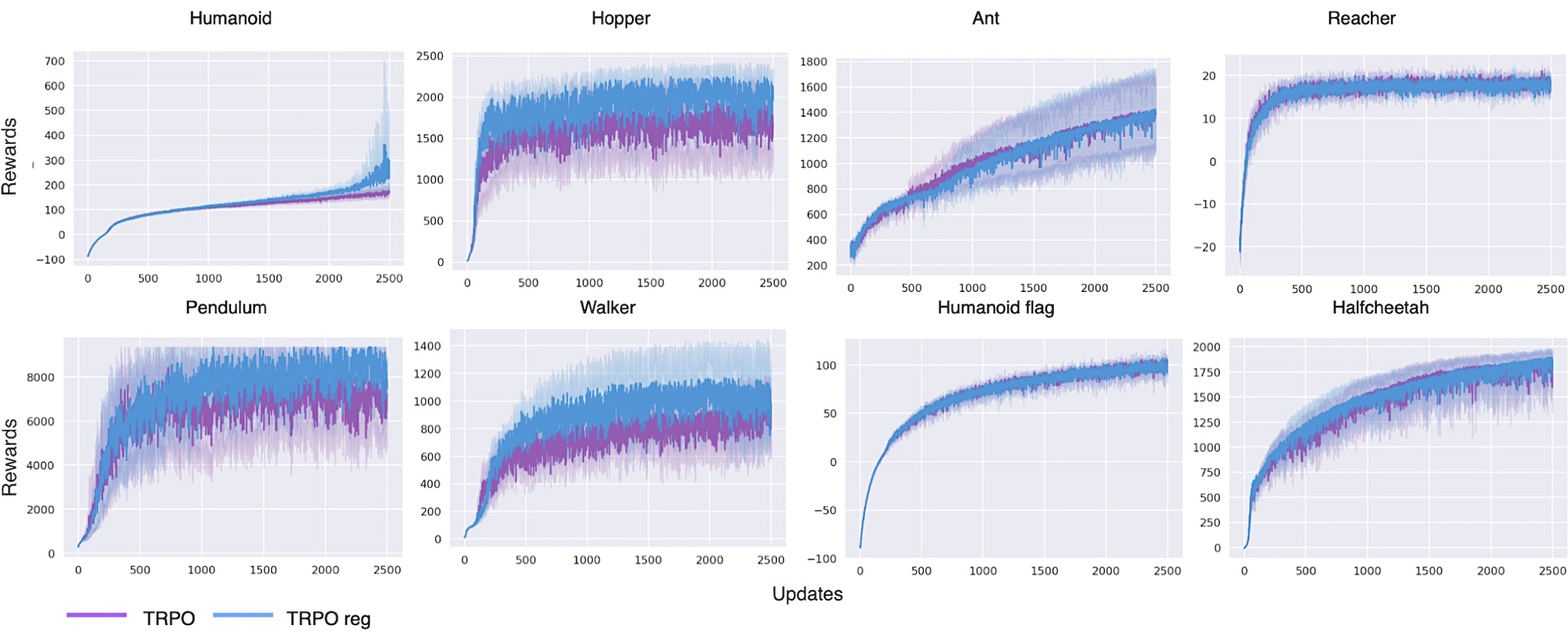}
    \vspace{-15pt}
    \caption{TRPO vs TRPO reg. Rewards and conditioning $\psi$, in PyBullet environments. Each curve is obtained by averaging results of 30 agents (10 for each seed)}
    \label{fig:3}
    \vspace{-12pt}
\end{figure*}

\begin{table*}[h!]
\begin{center}
 \begin{tabular}{c c c c c c c c c} 
 \hline
 Env & Humanoid & Hopper & Ant & Reacher & Pendulum & Walker & Humanoid Flag & HalfCheetah\\ [0.5ex] 
 \hline
 PPO  & 73.1 & 106.8 & 414.1 & -13.5 & 4534.9 & 55.1 & 43.6 & 9.9\\
 PPO reg & 73.2 & 18.6 & 294.8 & -7.4 & 4978.4 & 74.2 & 38.4 & 941.2\\
 \hline
 TRPO  & 165.6 & 1697.0 & 1376.8 & \textbf{18.2} & 6993.6 & 892.4 & \textbf{98.7} & 1795.3\\
 TRPO reg & \textbf{245.8} & \textbf{2001.2} & \textbf{1379.4} & 17.7 & \textbf{8213.8} & \textbf{1013.5} & \textbf{98.7} & \textbf{1843.3}\\
 \hline
\end{tabular}
\caption{Mean reward over the last 100 optimization steps for TRPO, PPO, PPO reg, and TRPO reg. The mean was computed over 3 random seeds and 10 agents for each seed using optimal policy hyperparameters}
\label{table:1}
\end{center}
\vspace{-15pt}
\end{table*}

To compute conditioning in RL agents, we feed two mini-batches at a time to the agent. The first batch consists of the real environment states $S_t$ at timestep $t$. The second batch consists of the same states but with some added disturbance $\delta$. Then we estimate how these batches affected the agent: $J_{t} = \frac{\|\pi_{\theta}(S_{t}) - \pi_{\theta}(S_{t}+\delta)\|}{\|\delta\|}.$
After this, we compute the value $\psi_t$ that characterizes how close $J_{t}$ is to the range $\lambda_{\max},\lambda_{\min}.$ These values approximately set the desirable range for model conditioning. We saved these parameters equal to the range defined previously for GANs ($1$ and $20$). 
\begin{equation}
\begin{split}
    \psi^{max}_{t}=\left(\max\left(J_{t},\lambda_{\max}\right)-\lambda_{\max}\right)^{2}, \\
    \psi^{min}_{t} =\left(\min\left(J_{t},\lambda_{\min}\right)-\lambda _{\min}\right)^{2}, \\
    \psi_{t} = \psi^{min}_{t} + \psi^{max}_{t}.
\end{split}
\end{equation}
More details are presented in Algorithm 1 in the Appendix. 

\textbf{Conditioning and Policy Performance:} To examine the relationship between policy and conditioning, we run PPO with different hyperparameters and random seeds on the four continuous control tasks Humanoid-v0, Hopper-v0, Ant-v0, and Reacher-v0. Through these trials, we try to examine whether ineffective policies are less conditioned. We use the standard PPO parameters as the optimal configuration and made three adjustments to those parameters to produce less effective policies.

In each configuration, we use the same minibatch size, the number of timesteps $T$, PPO epoch, policy learning rate, and $\eta$. Parameters that were modified are: GAE parameter, discount ($\gamma$), value function (VF) coefficient, VF learning rate, VF epochs~\cite{DBLP:journals/corr/SchulmanLMJA15, DBLP:journals/corr/SchulmanWDRK17}. The sets of hyperparameters are presented in Table 3 in the Appendix. We test each setting on 4 PyBullet environments with 3 random seeds and 10 agents for each seed. Results are presented in Figure~\ref{fig:1}. 

\textbf{Results Discussion:} Our experiments show that the conditioning has similar patterns with the number of received rewards. On the Humanoid task, we found that the most effective policy has the lowest conditioning. And furthermore, parameter 2 model drop of condition number corresponds to the moment of a sharp increase in rewards for the agent.

The connection between policy and Jacobian conditioning is not clearly evident in the Ant task. However, agents with parameters 2 and 3 that obtained smaller reward values are more distant from the dynamical isometry property. Furthermore, an interesting observation that is worth noting is that policies, which are well-performing and gain higher reward values at the end of the training, are better conditioned, often even from the first training steps. 

Because of environment dynamics, a linear relationship between the reward curves and the condition number is difficult to establish. However, in general, based on these experiments, a pattern can be observed: \textbf{a policy that receives fewer rewards has less optimal conditioning.} Also, turning back to the privileges that Dynamical Isometry provides for deep non-linear networks in classification and generation tasks too, we assume that if an agent is closer to Dynamical Isometry, it will allow forming a more stable and efficient policy.

\section{Conditioning Regularized Policy Optimization}
\begin{figure*}[h!]
    \centering
    \includegraphics[width=\linewidth]{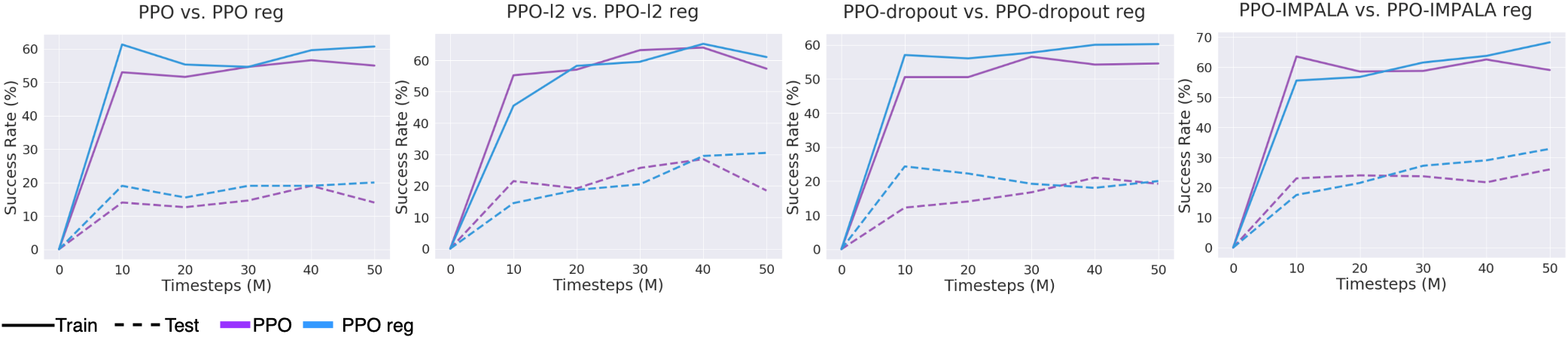}
    \vspace{-15pt}
    \caption{PPO and PPO with conditioning regularization. Success rate on CoinRun environment on train and test levels}
    \label{fig:4}
    \vspace{-5pt}
\end{figure*}

\begin{table*}[h!]
\begin{center}
 \begin{tabular}{c| c c|c c| c c| c c} 
 \hline
 Levels & PPO & PPO reg & PPO-l2 & PPO-l2 reg & PPO-D & PPO-D reg & PPO-IMP & PPO-IMP reg\\ [0.5ex] 
 \hline
 Seen  & 55 & \textbf{60.7} & 57.2 & \textbf{61} & 54.5 & \textbf{60.2} & 59 & \textbf{68.2}\\
 Unseen & 14 & \textbf{20} & 18.5 & \textbf{30.5} & 19.2 & \textbf{20} & 26 & \textbf{32.8}\\
 \hline
\end{tabular}
\caption{PPO and PPO with conditioning regularization. Success rate after 50M timesteps on CoinRun environment, on train(seen) and test(unseen) levels.}
\label{table:2}
\end{center}
\vspace{-15pt}
\end{table*}

In this section, we propose an algorithm that regularizes the condition number of the agent Jacobian. To regularize the policy we simply use the values of conditioning as a penalty. The example of regularized PPO presented below. We used the PPO algorithm and added a value of $\psi$ to the surrogate policy loss: 
\begin{equation}
\begin{split}
    L _ { t } ^ { C L I P + \psi + V F + S} ( \theta ) = \\
    \hat { \mathbb { E } } _ { t } \left[ L _ { t } ^ { C L I P } ( \theta ) +c _ { 1 }\psi - c _ { 2 } L _ { t } ^ { V F } ( \theta ) + c _ { 3 } S \left[ \pi _ { \theta } \right] \left( s _ { t } \right) \right],
    \end{split}
\end{equation}
where $ L ^ { C L I P }$ is PPO policy loss. $c _ { 1 }$ is coefficient for conditioning penalty,  $L_{ t }^{ V F }$ is a value loss $ \left (V _ {\theta} \left (s_{t} \right) - V_{t} ^ {\operatorname {targ} } \right) ^ {2} $ with coefficient $c_{ 2 }$, $S \left[ \pi _ { \theta } \right] \left( s _ { t } \right)$ is policy entropy for state $s_{t}$ multiplied by entropy coefficient $c_{ 3 }$. Conditioning penalty can be applied to other algorithms too, in our experiments we used it for TRPO as well. Condition value used for a penalty computing on the new policy on PPO and TRPO algorithm.

\textbf{Continuous Control Experiments:} We conduct experiments of the regularization technique on PPO and TRPO algorithms. We optimized 30 agents for each task (10 agents for 1 random seed) over 2500 updates (5 million timesteps) see Figure~\ref{fig:2},~\ref{fig:3}. We test algorithms on Humanoid-v0, Hopper-v0, Ant-v0, Reacher-v0, Double Inverted-Pendulum-v0, Humanoid-Flag-v0, Walker-v0, and Half-cheetah-v0 environments. 

In this test, the hyperparameters setting is equal to the optimal one, presented in PPO and TRPO literature~\cite{DBLP:journals/corr/SchulmanLMJA15, DBLP:journals/corr/SchulmanWDRK17} for continuous control tasks. For the TRPO algorithm, we also used mean conditioning of a trajectory as a penalty for surrogate policy loss. In our experiments both basic TRPO and regularized one show better results than PPO. The average rewards for the last 100 updates are shown in Table~\ref{table:1}. In all experiments model with name ``reg'' is conditioning regularized model. For experiments, we used the penalty multiplied by a coefficient $c _ { 1 }$ equal to 0.001. %As a result, we can see that often even soft conditioning regularization can produce better policy performance

\textbf{Generalization Experiments:} Our continual learning problem was set without explicitly separated training and testing stages. In generalization experiments, we trained models on the fixed large-scale set of 500 levels of CoinRun~\cite{DBLP:conf/icml/CobbeKHKS19} and tested on unseen levels. In this experiment, we run PPO with $l2$ and Dropout~\cite{DBLP:journals/jmlr/SrivastavaHKSS14} regularizations, then we run the same methods but with additional conditioning penalty. For this experiment, we use ``NatureCNNs'' architecture proposed for tests in~\cite{DBLP:conf/icml/CobbeKHKS19}. Also, we tested the PPO method without $l2$ and dropout regularization but based on IMPALA~(IMP)~\cite{DBLP:conf/icml/EspeholtSMSMWDF18} architecture.  

We noticed a high variance in scores during tests. Due to that, at evaluation, we increase the number of repeats form 5 as it was used in~\cite{DBLP:conf/iclr/LeeLSL20} to 20. We trained models over 50M timesteps, but only on one random seed, all other settings were equal to~\cite{DBLP:conf/iclr/LeeLSL20} (Section 4.2). Results are presented in Figure~\ref{fig:4} and Table~\ref{table:2}. Our method outperforms PPO in all 4 training scenarios.

\section{Discussion and Future Work} 
In this work, we propose a simple and computationally inexpensive optimization method for Deep RL. We adapted a technique called Jacobian Clamping to approximately estimate conditioning of the agent. We tested our approach on the PyBullet and CoinRun domains. In our opinion, extending RL algorithms by conditioning regularization is a promising research direction. Condition number can provide important information about the policy, such as the correctness of hyperparameters or stability.

However, our work is still in progress. To study the role of conditioning for the generalization problem more thoroughly, we plan to conduct a test on the CoinRun environment with more timesteps and random seeds. Our experiments show that different architectures conditioning regularization produces various results. We plan to test conditioning contribution to other architectures too and run them on the environments like DeepMind Lab~\cite{DBLP:journals/corr/BeattieLTWWKLGV16}. Also, we plan to compare conditioning regularization with other methods such as information bottleneck~\cite{DBLP:conf/iclr/GoyalISALBBL19, DBLP:conf/iclr/GalashovJHTSDCT19, DBLP:conf/nips/IglCLTZDH19}. Estimating squared singular values of the agent Jacobian matrix using SVD would be a very interesting experiment to examine the role of Dynamical Isometry in RL agents.
%The perspective direction of the research is the development of techniques that make the clip parameter more sensitive to the difference between old and new policy conditioning. Another direction is the application of the proposed technique to more complex environments and tasks. Finally, an important direction is the use of this technique in state-of-the-art methods based on PPO.

\bibliographystyle{icml2020}
\bibliography{bib.bib}

\end{document}